%% file: root.tex
\documentclass[letterpaper, 10 pt, conference]{ieeeconf}  

\IEEEoverridecommandlockouts                              

\usepackage{graphics} 
\usepackage{epsfig} 
\usepackage{mathptmx} 
\usepackage{times} 
\usepackage{amsmath} 
\usepackage{amssymb}  
\usepackage{algorithm}
\usepackage{algpseudocode}          
\usepackage{booktabs}
\usepackage{multirow} 
\usepackage{dsfont}
\usepackage{amsfonts}
\usepackage{newtxtext} 
\usepackage{xcolor}
\usepackage{bbm}                       
\usepackage{textcomp}
\usepackage{hyperref}
\newcommand{\cmark}{$\checkmark$}
\newcommand{\xmark}{$\times$}
\title{\LARGE \bf
TactileReflex: Noise-Statistics-Driven Vision-Tactile Reflex Control for Force-Sensitive Manipulation
}

\author{%
Ziyan Feng$^{1}$,
Yulong Fu$^{1}$,
Zheng Li$^{1}$,
Yuxin He$^{1}$,
Jieji Ren$^{2}$,
Yudong Zhong$^{1}$,\\
Lujia Wang$^{1}$,
Jinni Zhou$^{1}$,
Qiang Nie$^{1,*}$%
\thanks{$^{1}$Thrust of Robotics and Autonomous Systems, The Hong Kong University of Science and Technology (Guangzhou), Guangzhou, China.}%
\thanks{$^{2}$ School of Mechanical Engineering, Shanghai Jiao Tong University, Shanghai, China.}%
\thanks{$^{*}$Corresponding author: Qiang Nie, \texttt{qiangnie@hkust-gz.edu.cn}.}%
}
\begin{document}

\maketitle
\thispagestyle{empty}
\pagestyle{empty}

\begin{abstract}
Manipulating fragile deformable containers, such as disposable
plastic cups filled with liquid, demands real-time grip-force
adaptation within an extremely narrow force margin: insufficient force
causes slip, while excessive force irreversibly deforms the thin wall.
Existing approaches struggle to achieve such force-sensitive manipulation tasks.
We propose a \emph{noise-statistics-based calibration-driven reflex
control paradigm with vision-based tactile sensing}: by analyzing the
sensor's intrinsic noise characteristics
(via a brief static-hold-and-unload protocol), we directly derive
all controller thresholds, eliminating external force
calibration, trial-and-error manual tuning, or material-specific physical models.
Instantiating this paradigm, we present \textit{TactileReflex}, a
three-channel closed-loop controller that extracts three image-level
proxies, shear intensity ($S_y$), contact intensity ($F_n$),
and center of pressure ($\boldsymbol{C}$), from dual visuo-tactile sensors
and drives prioritized reflex channels at ${\sim}$12\,Hz for slip suppression,
weight-adaptive release, and force protection.
Each channel closes the loop directly on its proxy via noise-derived thresholds.
Ablation demonstrates that only the full three-channel system is able to prevent irreversible container deformation (5/5 success vs.\ at most 1/5 for partial configurations).
In a dynamic pouring task, fixed-effort baselines fail in all
10 attempts due to \textbf{pose drift}, while TactileReflex achieves
9/10 success across two water volumes.
As a self-contained and interpretable controller, TactileReflex can
serve as a plug-and-play safety layer beneath high-level manipulation
pipelines, including \textbf{haptic-free} VR teleoperation and vision-language-action (VLA) policies.

\textbf{Project page:} \href{https://shayfeng.github.io/TactileReflex/}{https://shayfeng.github.io/TactileReflex/}
\end{abstract}

\section{INTRODUCTION}
\label{sec:intro}

Manipulating deformable containers (soft plastic cups, thin-walled
bottles, flexible pouches) is a daily task for humans yet remains
challenging for robots.
Consider a commonplace disposable plastic cup: it weighs only
3.5--5\,g, its wall is just 0.3--0.5\,mm thick, it is transparent
(offering minimal visual contrast for external cameras), and the
margin between the minimum force needed to prevent slip and the force
that permanently crushes the cup is remarkably narrow.
Now fill it with water and pour: the shifting liquid continuously
redistributes the gravitational load along the gripper fingers,
demanding real-time effort modulation that fixed-effort or open-loop
grasping cannot achieve.
The core difficulty is that grasp stability and object safety are
tightly coupled: insufficient effort leads to micro-slip and drop,
while only slightly more force causes irreversible deformation.
A practical grasp controller must therefore
(1)~detect and suppress incipient slip in real time,
(2)~reduce effort when the carried load decreases to prevent
over-gripping, and
(3)~enforce a hard safety limit on contact force.
The central challenge is ensuring grasp stability while preventing
irreversible deformation during dynamic tasks.

\textbf{Limitations of existing approaches.}\;
Force/torque sensors can provide continuous feedback for adaptive grasping, but typically require per-sensor calibration and are sensitive to mounting/compliance effects~\cite{chavez2019insitu,ati_ft_faq2020}.
Vision-based tactile sensors~\cite{yuan2017gelsight,lambeta2020digit} offer rich contact imagery that can address the \emph{perception} side of this challenge, yet are typically used for binary slip classification~\cite{dong2019gelsight_slip} or as inputs to end-to-end learned policies~\cite{tafvla2024}, neither of which constitutes an effective control framework.
Classification outputs only a discrete decision without specifying \emph{how much} to adjust the grasp,
while end-to-end policies, without interpretable intermediates, cannot impose transparent safety bounds.

\textbf{Our approach.}\;
To bridge this gap, we propose a \emph{noise-statistics-based calibration-driven reflex
control paradigm} for vision-based tactile sensing.
A brief (${\sim}$2\,min) static-hold-and-unload calibration captures the
sensor's intrinsic noise profile, from which percentile quantiles
(e.g., $P_{95}$, $P_{99.9}$) automatically set all
controller thresholds---eliminating trial-and-error manual tuning and
any external force reference.
This paradigm also clarifies how thresholds transfer across objects:
some thresholds are robust to material/contact geometry,
while the rest are re-derived by repeating the calibration on each new target, without modifying the controller.

Instantiating this paradigm, we extract three interpretable image-level
proxies---shear intensity $S_y$, contact intensity $F_n$, and center of
pressure $\boldsymbol{C}$---and close the loop \emph{directly},
without intermediate classification or explicit force estimation.
Embedding these signals in a prioritized reflex architecture yields a
controller functionally analogous to the human grasp reflex~\cite{johansson1984}:
fast, involuntary grip-force adjustments that stabilize the grasp
independent of higher-level policies.

\textbf{Contributions.}
\begin{itemize}
\item A \emph{noise-statistics-based calibration-driven reflex control paradigm with
      vision-based tactile sensing} that derives
      all controller thresholds automatically from the sensor's intrinsic
      noise statistics, requiring no external force
      sensor, no trial-and-error manual parameter tuning, and no material-specific physical models.
\item \textit{TactileReflex}, an instantiation of this paradigm as a
      three-channel closed-loop controller (anti-slip\,/\,weight-adaptive
      release\,/\,force-protect) that operates at ${\sim}$12\,Hz using
      only tactile images, with experimentally validated convergence behavior for
      each channel and robustness to severe left--right sensor asymmetry.
\item Systematic real-robot validation demonstrating:
      (i) cross-material calibration achieves 100\% true-positive and
      0\% false-positive rates for slip detection;
      (ii) ablation confirms that only the full three-channel system 
      is able to prevent irreversible container deformation (5/5 success vs.\ max 1/5);
      and (iii) a dynamic pouring task where fixed-effort baselines
      universally fail (0/10) while TactileReflex succeeds (9/10).
\end{itemize}

\input{related_work}

\input{system}

\input{method}

\input{experiments}

\input{discussion}


\bibliographystyle{IEEEtran}
\bibliography{IEEEabrv,reference}

\end{document}

%% file: related_work.tex
\section{RELATED WORK}
\label{sec:related}

Gel-based optical tactile sensors (e.g., GelSight~\cite{yuan2017gelsight}, DIGIT~\cite{lambeta2020digit}, GelSlim~\cite{donlon2018gelslim}, MC-Tac~\cite{ren2023mctac}) provide dense contact geometry without per-taxel calibration. While typically exploited for high-level perception~\cite{li2014localization, bauza2023tac2pose, calandra2018more} or binary slip classification via analytical~\cite{dong2019gelsight_slip, james2018slip, su2015force} and learned~\cite{li2018slip, calandra2017feeling} methods, these paradigms often introduce perception--action latency. A smaller body of work explores tighter tactile servoing~\cite{veiga2015stabilizing, hogan2018tactile}, yet most evaluate detection accuracy in isolation without demonstrating closed-loop grasp stability under continuous dynamic loading. Building on this ``control-first'' philosophy, we bypass high-level feature extraction. Instead, we close the loop directly using three minimal image-level proxies ($S_y, F_n, \boldsymbol{C}$) and a noise-statistics-based calibration, eliminating the need for task-specific training or discrete slip detection.

Classical force-adaptive grasping relies on F/T sensors~\cite{romano2011human, dafle2014extrinsic}, which are sensitive to mounting compliance and challenging to calibrate for continuously deforming objects. Conversely, recent deformable-manipulation learning frameworks and benchmarks~\cite{seita2021learning_deformable, lin2020softgym} advance data-driven control, while Vision-Language-Action (VLA) models (e.g., RT-2~\cite{brohan2023rt2}, $\pi_0$~\cite{black2024pi0}) produce end-to-end commands but lack built-in, hard contact-force guarantees. Even tactile-integrated VLAs like TaF-VLA~\cite{tafvla2024} rely on the learned policy for force regulation. To bridge this gap, we design TactileReflex as a complementary, low-level reflex layer~\cite{liu2024safety_layer}. Analogous to human spinal reflexes~\cite{johansson1984}, it uses tactile proxies as surrogate force signals to enforce explicit safety bounds. By intercepting gripper commands, it transparently enables grasp stability at ${\sim}$12\,Hz beneath any high-level semantic planner.

%% file: system.tex
\section{SYSTEM OVERVIEW}
\label{sec:system}

\begin{figure*}[t]
\vspace{0.2cm}
  \centering
  \begin{minipage}[c]{0.533\textwidth}
    \centering
    \includegraphics[width=\linewidth,height=2.8in,keepaspectratio]{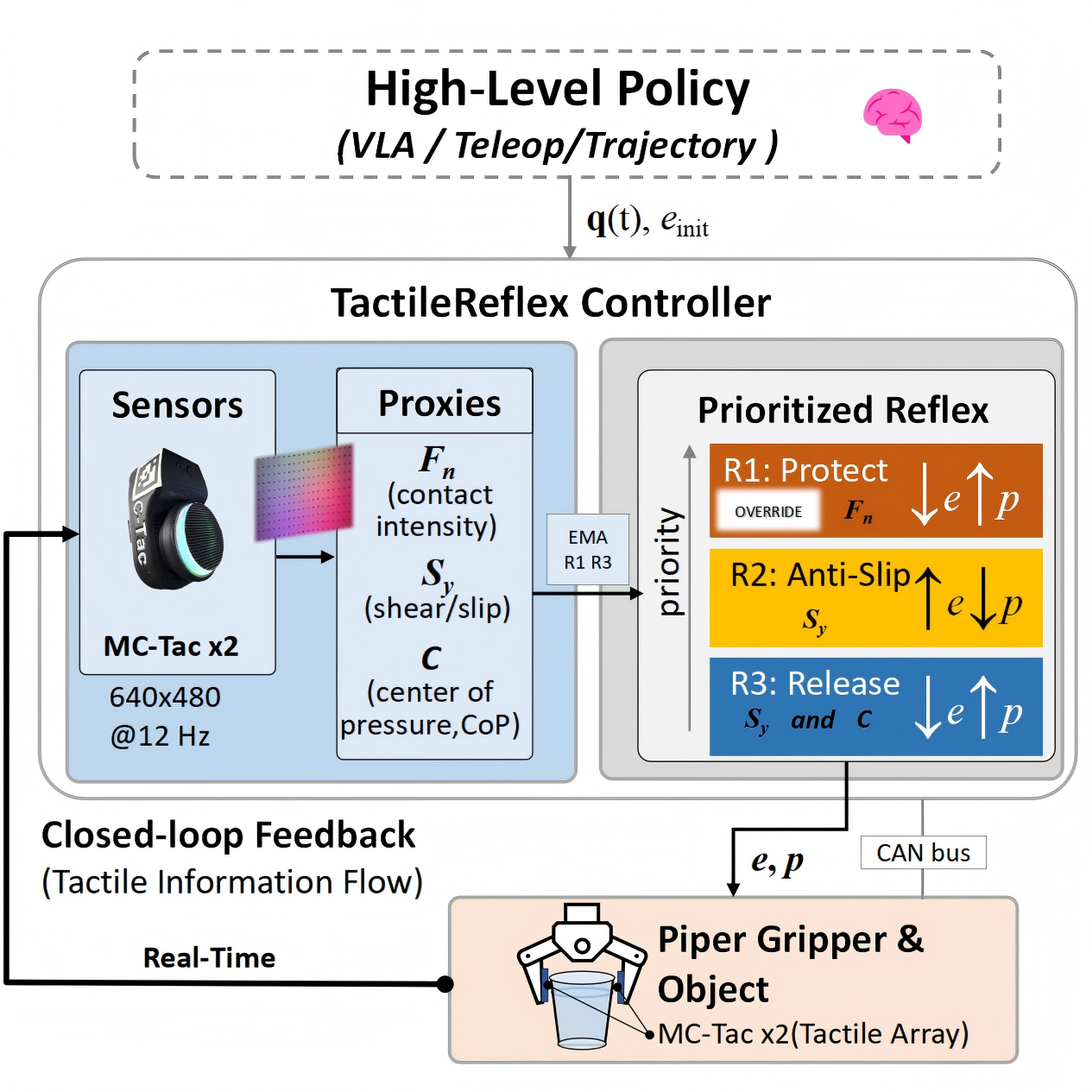}
    \centerline{\small (a) System architecture}
  \end{minipage}
  \hfill
  \begin{minipage}[c]{0.46\textwidth}
    \centering
    \begin{minipage}[t]{0.6\linewidth}
      \centering
      \includegraphics[width=\linewidth]{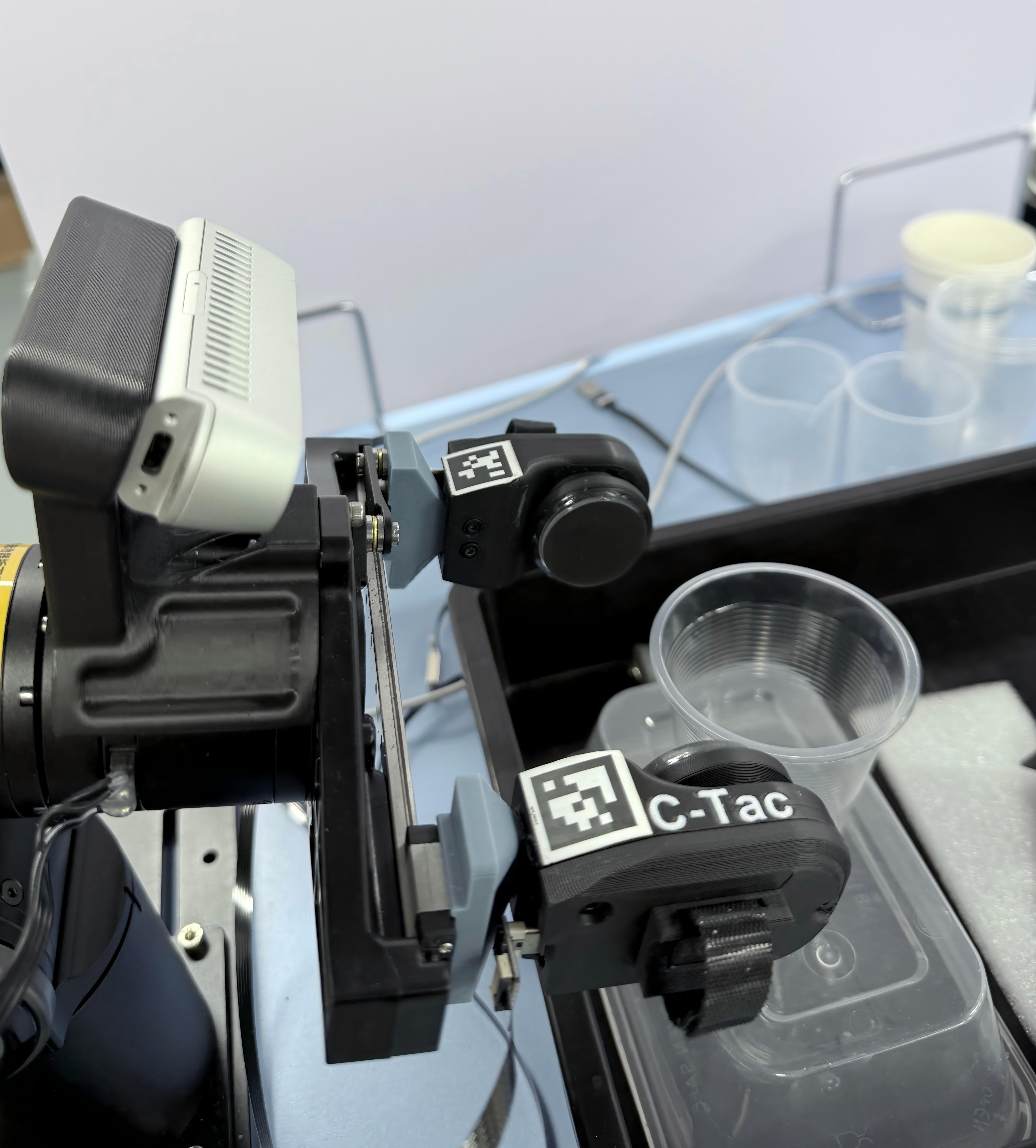}
      \centerline{\small (b) Hardware and experiment setup}
    \end{minipage}
    \vspace{12pt}
    \begin{minipage}[t]{0.9\linewidth}
      \centering
      \includegraphics[width=\linewidth]{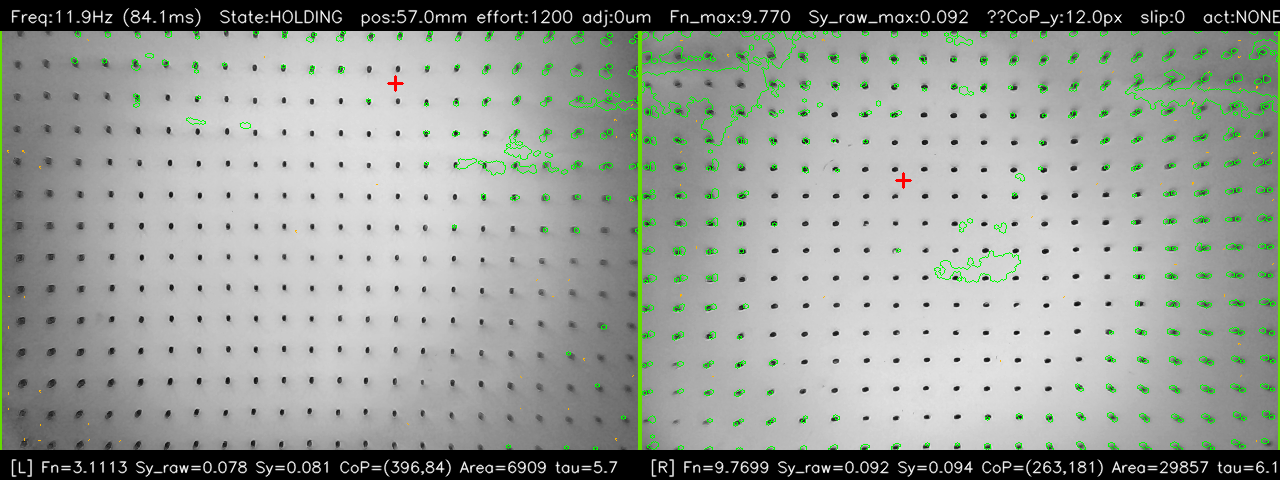}
      \centerline{\small (c) Processed left and right tactile images}
    \end{minipage}
  \end{minipage}
  \caption{System overview.
  (a)~The TactileReflex layer sits beneath a high-level policy (VLA/teleop/trajectory), which provides the arm joint trajectory $\mathbf{q}(t)$ and a nominal grasp effort $e_{\mathrm{init}}$.
  It extracts three image-level proxies
  (shear intensity $S_y$, contact intensity $F_n$, center of pressure $\boldsymbol{C}$)
  from dual MC-Tac sensors and feeds
  them into three prioritized reflex channels (anti-slip, release, force-protect)
  that jointly adjust gripper effort~$e$ and stroke position~$p$.
  The anti-slip channel tightens the grasp by increasing $e$ while decreasing $p$,
  whereas the release and force-protect channels loosen it by decreasing $e$ while increasing $p$;
  force-protect can override the other two channels.
  (b)~Hardware and experiment setup with MC-Tac sensors on the Piper gripper.
  (c)~Processed tactile images from the left and right finger sensors, showing proxy extraction and real-time monitoring.}
  \label{fig:system}
\end{figure*}

Fig.~\ref{fig:system} shows the overall architecture and hardware.
The TactileReflex layer sits between a high-level policy
(trajectory replay, teleoperation, or a learned planner) and the
gripper hardware, intercepting gripper commands and applying
real-time tactile corrections.

\subsection{Hardware Platform}

Our platform consists of a single-arm robot equipped with a Piper
parallel gripper.
Two MC-Tac vision-based tactile sensors~\cite{ren2023mctac}
are mounted on opposing fingertips via custom 3D-printed adapters,
each capturing $640{\times}480$ tactile images.
To ensure measurement consistency, camera exposure and white balance
are fixed during operation.
\subsection{Software Architecture}

The control loop is implemented in a single Python process
running on a standard workstation.
Each cycle performs the following steps sequentially:
(i)~capture a synchronized pair of tactile images,
(ii)~compute per-sensor proxies ($F_n$, $S_y$, $\boldsymbol{C}$) and EMA updates
    (Sec.~\ref{sec:proxies}),
(iii)~evaluate the reflex state machine
    (Sec.~\ref{sec:controller}), and
(iv)~send the resulting effort and position commands to the gripper.
The system operates at approximately 12\,Hz, sufficient for real-time reflex control.

\subsection{Applicability as a Transparent Safety Layer}
\label{sec:vr_teleop}

Because TactileReflex operates entirely at the gripper level and
exposes only a standard effort/position interface, it can be inserted
beneath any high-level manipulation pipeline---fixed trajectories,
learned policies, or teleoperation---without modifying the upstream
planner.
A representative use case is haptic-free VR teleoperation: standard
controllers (e.g., Meta Quest~3) lack precise force feedback, making
demonstrations on fragile objects prone to crushing.
With TactileReflex enforcing contact-safety bounds locally, operators
can collect high-quality dynamic manipulation data (e.g., pouring
trajectories for deformable containers) without attending to
low-level force constraints, thereby lowering the barrier to scalable demonstration
collection for robot learning.

%% file: method.tex
\section{METHOD}
\label{sec:method}

\subsection{Image-Level Tactile Processing}
\label{sec:proxies}

Each MC-Tac sensor captures $640{\times}480$ grayscale images.
Let $I_t$ denote the image at time~$t$ and let $I_{\mathrm{ref}}$ be a
no-contact reference (computed during initialization and periodically
updated during non-contact phases).
Let $u = (x, y)$ denote pixel coordinates, with $H{=}480$ and $W{=}640$
being image height and width.
The per-pixel difference is
\begin{equation}
  \Delta I_t(u)=\bigl|I_t(u)-I_{\mathrm{ref}}(u)\bigr|.
  \label{eq:deltaI}
\end{equation}

Thresholding $\Delta I_t$ at noise floor $\tau$ ($P_{98}$, the 98th percentile of pixel noise;
Sec.~\ref{sec:calib}) followed by morphological opening yields a binary contact
mask~$\hat{B}_t$ and the contact support $\mathcal{U}_t = \{u \mid \hat{B}_t(u){=}1\}$;
a frame is considered no-contact if $|\mathcal{U}_t| < N_{\min}$,
where $N_{\min}$ denotes the minimum contact pixel count.
The tactile weight is
\begin{equation}
  W_t(u)=
  \begin{cases}
    \Delta I_t(u)^{\gamma}, & u \in \mathcal{U}_t,\\
    0, & \text{otherwise},
  \end{cases}
  \label{eq:weight}
\end{equation}
where $\gamma > 0$ is a sensitivity exponent.
Grayscale processing avoids spectral sensitivity to lighting changes;
$\Delta I_t$ further cancels global brightness shifts and $\gamma{>}1$
suppresses residual noise, yielding robustness to ambient illumination.

We compute three image-level proxies on each sensor ($s \in \{L, R\}$).

\textbf{Normal-contact proxy.}
\begin{equation}
  F_n^{\mathrm{raw},s}(t)=\frac{1}{HW}\sum_{u} W_t^s(u),
  \label{eq:fn_raw}
\end{equation}
where the superscript $\mathrm{raw}$ denotes the unfiltered
instantaneous reading (before the EMA smoothing introduced below)
and $s \in \{L, R\}$ indexes the left and right sensor.
$F_n$ is a dimensionless contact-intensity proxy, not an absolute
force in Newtons.

\textbf{Center of pressure (CoP).}
\begin{equation}
  \boldsymbol{C}_t^s=\frac{\sum_{u} W_t^s(u)\,u}{\sum_{u} W_t^s(u)}.
  \label{eq:cop}
\end{equation}
Let $C_{y,t}^s$ denote the vertical (gravity-aligned) component of $\boldsymbol{C}_t^s$. The mean vertical center of pressure across both sensors is:
\begin{equation}
  \hat{C}_{y,t} = \tfrac{1}{2}\bigl(C_{y,t}^{L} + C_{y,t}^{R}\bigr).
  \label{eq:cop_avg}
\end{equation}
This is computed only when both sensors detect valid contact
($|\mathcal{U}_t^{L}| \ge N_{\min}$ and $|\mathcal{U}_t^{R}| \ge N_{\min}$).

\textbf{Shear/slip proxy.}
Using Farneback optical flow~\cite{farneback2003}
$\mathbf{v}_t^s(u) = (v_x, v_y)$ computed between consecutive frames:
$\mathbf{v}_t^s = \mathrm{Farneback}(I_{t-1}^s, I_t^s)$, where
$\mathbf{v}_t^s(u)$ is the optical flow vector at pixel $u$:
\begin{equation}
  S_y^{\mathrm{raw},s}(t)=\operatorname{median}_{u \in \mathcal{U}_t^s}\!\bigl|v_{y,t}^s(u)\bigr|.
  \label{eq:sy_raw}
\end{equation}
The median is area-invariant: it remains robust even under severe left-right contact area asymmetry, maintaining a consistent noise floor (Sec.~\ref{sec:exp_crossmat}), making the slip threshold $\theta_s$ (Sec.~\ref{sec:calib}) robust to contact geometry.

Selected channels undergo exponential moving average (EMA) smoothing:
$\tilde{x}_t = \alpha x_t^{\mathrm{raw}} + (1{-}\alpha)\tilde{x}_{t-1}$,
and are aggregated across sensors via element-wise maximum
($\tilde{S}_y^{\max}$, $\tilde{F}_n^{\max}$).
As Table~\ref{tab:raw_ema} shows, anti-slip uses the raw signal to
maintain responsiveness, while release and protection employ
EMA-filtered signals to mitigate high-frequency noise.

\begin{table}[t]
  \centering
  \caption{Signal processing pipeline for each reflex channel.
  Thresholds: $\theta_s$ (slip), $\theta_q$ (quiet shear), $F_{\mathrm{lim}}$ (force limit).
  $P_{95}$/$P_{99.9}$: 95th/99.9th percentiles of static-hold noise.}
  \label{tab:raw_ema}
  \begin{tabular}{lllll}
    \toprule
    Channel & Signal & Filtering & Threshold & Calibration Basis \\
    \midrule
    Anti-slip & $S_y^{\mathrm{raw},s}$ & None & $\theta_s$ & Static/Slip $S_y^{\mathrm{raw}}$ \\
    Release   & $\tilde{S}_y^{\max}$   & EMA  & $\theta_q$ & $P_{95}(S_y^{\mathrm{raw}})$ \\
    Protection & $\tilde{F}_n^{\max}$  & EMA  & $F_{\mathrm{lim}}$ & $P_{99.9}(F_n^{\mathrm{raw}})$ \\
    \bottomrule
  \end{tabular}
\end{table}
\subsection{Three-Channel Reflex Control}
\label{sec:controller}

The system operates in three sequential states:
\textsc{Idle} (gripper open),
\textsc{Closing} (stroke decreasing at fixed effort $e_{\mathrm{init}}$
until $\min(F_n^{\mathrm{raw},L}, F_n^{\mathrm{raw},R}) \ge F_{\mathrm{stop}}$),
where $F_{\mathrm{stop}}$ is the grasp-stop force threshold,
and \textsc{Holding} (reflex control active).
The three reflex channels described below are active only
during \textsc{Holding}.

\textbf{Control variables and channel priority.}\;
We jointly control gripper effort $e_t \in [e_{\mathrm{init}}, e_{\max}]$
and stroke position $p_t \in [p_{\min}, p_{\mathrm{hold}}]$,
where $e_{\mathrm{init}}$ is the initial holding effort,
$e_{\max}$ is the maximum effort limit,
$p_{\min}$ is the minimum stroke position,
and $p_{\mathrm{hold}}$ is the position when stable holding was first established.
Each reflex channel adjusts both variables in a synergistic manner.
R2 (anti-slip) and R3 (release) are mutually exclusive; at most one fires per cycle.
R1 (force-protect) is evaluated independently after R2/R3 and can override their effects;
when R1 co-triggers with R2, the net effort change is
$+\Delta e_{\mathrm{slip}} - \Delta e_{\mathrm{prot}}$,
always yielding a decrease since $\Delta e_{\mathrm{prot}} > \Delta e_{\mathrm{slip}}$
by design.

\subsubsection{(R1) Force-Protect (Highest Priority)}
If $\tilde{F}_n^{\max} > F_{\mathrm{lim}}$:
\begin{align}
  e_{t+1} &= \max\!\bigl(e_t' - \Delta e_{\mathrm{prot}},\; e_{\mathrm{init}}\bigr),
  \label{eq:protect_e} \\
  p_{t+1} &= \min\!\bigl(p_t' + \Delta p_{\mathrm{prot}},\; p_{\mathrm{hold}}\bigr),
  \label{eq:protect_p}
\end{align}
where $e_t', p_t'$ denote effort and position after any R2/R3 action in the same cycle.

\subsubsection{(R2) Anti-Slip}
Slip is detected per-sensor with contact gating:
\begin{equation}
  \mathrm{slip}_s = \mathds{1}\!\bigl[|\mathcal{U}_t^s| \ge N_{\min}\bigr]
  \;\cdot\; \mathds{1}\!\bigl[S_y^{\mathrm{raw},s} \ge \theta_s\bigr],
  \quad s \in \{L, R\}.
  \label{eq:slip_detect}
\end{equation}
If $\mathrm{slip}_L \lor \mathrm{slip}_R$:
\begin{align}
  e_{t+1} &= \min\!\bigl(e_t + \Delta e_{\mathrm{slip}},\; e_{\max}\bigr),
  \label{eq:antis_e} \\
  p_{t+1} &= \max\!\bigl(p_t - \Delta p_{\mathrm{slip}},\; p_{\min}\bigr),
  \label{eq:antis_p}
\end{align}
subject to $\textstyle\sum \Delta p_{\mathrm{slip}} \le \Delta p_{\max}$,
where $\Delta p_{\max}$ is the maximum total stroke reduction to avoid excessive gripper tightening and irreversible container deformation.
The contact gate prevents false triggers from non-contacting sensors.
Using raw $S_y$ preserves single-frame responsiveness.

\subsubsection{(R3) Weight-Adaptive Release}
We track the CoP reference $\bar{C}_y$ as a slowly varying
baseline. Let $\Delta C_{y,t} = \hat{C}_{y,t} - \bar{C}_y$ denote the
CoP shift along the gravity-aligned axis.

Release fires only when three preconditions hold:
(i)~dual contact on both sensors,
(ii)~CoP reference $\bar{C}_y$ initialized and tracked,
(iii)~quiet shear $\tilde{S}_y^{\max} < \theta_q$.

The CoP reference $\bar{C}_y$ is maintained at \emph{two rates}:
\begin{itemize}
\item \emph{Background} (every holding frame):
  $\bar{C}_y \!\leftarrow\! (1{-}\alpha_{\mathrm{bg}})\bar{C}_y
    + \alpha_{\mathrm{bg}}\hat{C}_{y,t}$,
  $\alpha_{\mathrm{bg}}{=}0.02$,
  compensating slow sensor drift;
\item \emph{Release} (only when R3 fires):
  $\bar{C}_y \!\leftarrow\! (1{-}\alpha_{\mathrm{rel}})\bar{C}_y
    + \alpha_{\mathrm{rel}}\hat{C}_{y,t}$,
  $\alpha_{\mathrm{rel}}{=}0.3$,
  enabling convergence within 3--4 frames.
\end{itemize}

If all preconditions hold and $\Delta C_{y,t} < \theta_c$:
\begin{align}
  e_{t+1} &= \max\!\bigl(e_t - \Delta e_{\mathrm{rel}},\; e_{\mathrm{init}}\bigr),
  \label{eq:release_e} \\
  p_{t+1} &= \min\!\bigl(p_t + \Delta p_{\mathrm{rel}},\; p_{\mathrm{hold}}\bigr).
  \label{eq:release_p}
\end{align}

\textbf{Closed-loop property.}\;
The fast CoP tracking drives $\bar{C}_y$ toward $\hat{C}_{y,t}$,
shrinking $|\Delta C_{y,t}|$ each frame until $\Delta C_{y,t} > \theta_c$
and release halts automatically.

Each channel applies fixed per-cycle increments
$\Delta e_{\mathrm{slip}}, \Delta e_{\mathrm{prot}}, \Delta e_{\mathrm{rel}}$
to effort and
$\Delta p_{\mathrm{slip}}, \Delta p_{\mathrm{prot}}, \Delta p_{\mathrm{rel}}$
to stroke position, with the design constraint
$\Delta e_{\mathrm{prot}} > \Delta e_{\mathrm{slip}}$ so that
force-protect always yields a net effort decrease when co-triggering
with anti-slip.
Algorithm~\ref{alg:reflex} summarizes the per-cycle control loop.

\begin{figure}[t]
\begin{algorithm}[H]
\caption{Tactile Reflex Control (per cycle, \textsc{Holding} state)}
\label{alg:reflex}
\begin{algorithmic}[1]
\Require Tactile images $I_t^L, I_t^R$; current effort $e$, position $p$
\State Compute proxies $F_n^{\mathrm{raw},s},\, S_y^{\mathrm{raw},s},\, \boldsymbol{C}^s$ for $s \in \{L,R\}$
\State Update EMA: $\tilde{F}_n^s,\, \tilde{S}_y^s$
\State Background CoP tracking:
  $\bar{C}_y \!\leftarrow\! (1{-}\alpha_{\mathrm{bg}})\bar{C}_y
    + \alpha_{\mathrm{bg}}\hat{C}_{y,t}$
\State $\mathrm{slip}_s \leftarrow 
\mathds{1}[|\mathcal{U}_t^s|\ge N_{\min}]\cdot 
\mathds{1}[S_y^{\mathrm{raw},s}\ge\theta_s]$ 
for $s \in \{L,R\}$
\If{$\mathrm{slip}_L \lor \mathrm{slip}_R$}
  \State $e \leftarrow \min(e + \Delta e_{\mathrm{slip}},\, e_{\max})$
  \State $p \leftarrow \max(p - \Delta p_{\mathrm{slip}},\, p_{\min})$
\ElsIf{dual contact $\land\; \Delta C_{y,t} < \theta_c
  \;\land\; \tilde{S}_y^{\max} < \theta_q$}
  \State $e \leftarrow \max(e - \Delta e_{\mathrm{rel}},\, e_{\mathrm{init}})$
  \State $p \leftarrow \min(p + \Delta p_{\mathrm{rel}},\, p_{\mathrm{hold}})$
  \State $\bar{C}_y \!\leftarrow\!
    (1{-}\alpha_{\mathrm{rel}})\bar{C}_y
    + \alpha_{\mathrm{rel}}\hat{C}_{y,t}$
\EndIf
\If{$\tilde{F}_n^{\max} > F_{\mathrm{lim}}$}
  \State $e \leftarrow \max(e - \Delta e_{\mathrm{prot}},\, e_{\mathrm{init}})$
  \State $p \leftarrow \min(p + \Delta p_{\mathrm{prot}},\, p_{\mathrm{hold}})$
\EndIf
\State Send $(p,\, e)$ to gripper
\end{algorithmic}
\end{algorithm}
\end{figure}

\subsection{Noise-Statistics-Based Threshold Calibration}
\label{sec:calib}
All controller thresholds are derived from a single, brief
(${\sim}$2\,min) static-hold-and-unload calibration that
characterizes the sensor's intrinsic noise profile, instantiating
the noise-statistics-based paradigm introduced in
Sec.~\ref{sec:intro}.
Percentile quantiles of the recorded proxy signals serve as
principled decision boundaries:
the noise floor $\tau = P_{98}(\Delta I)$,
the quiet-shear threshold $\theta_q = P_{95}(S_y^{\mathrm{raw}})$,
the force protection limit
$F_{\mathrm{lim}} = P_{99.9}(F_n^{\mathrm{raw}})$,
and the slip threshold $\theta_s$ is placed by a conservative rule
just above the noise ceiling to capture the weakest observed slip
(100\% TPR, 0\% FPR on both test materials),
where TPR = correctly detected slips\,/\,all actual slips and
FPR = false alarms\,/\,all non-slip samples.
The CoP threshold $\theta_c$ is derived from manual load-decrease
events during the same protocol.

As Table~\ref{tab:thresholds} summarizes, the resulting parameters
separate into two groups:
\emph{universal or near-universal} parameters
($\theta_s$, $\theta_q$, $\gamma$, $\alpha$) whose values remain
constant or nearly so across different objects, and
\emph{material-dependent} parameters
($F_{\mathrm{lim}}$, $\theta_c$, $F_{\mathrm{stop}}$) that are
automatically re-derived by repeating the same brief protocol on
each new target object.
This two-tier structure ensures that adapting the controller to an
unseen object requires only re-running the calibration---no manual
parameter search, no retraining, and no structural change.
Because all thresholds originate from raw-signal statistics, which
are inherently noisier than the EMA-filtered signals used at runtime,
triggering is conservative by construction.

\begin{table}[t]
\vspace{0.2cm}
  \centering
  \caption{Threshold values and their generality across materials.}
  \label{tab:thresholds}
  \begin{tabular}{llccl}
    \toprule
    Parameter & Symbol & Soft & Hard & Generality \\
    \midrule
    Slip threshold  & $\theta_s$ & 0.20 & 0.20
      & \cmark~univ. \\
    Quiet threshold & $\theta_q$ & 0.12 & 0.10
      & $\approx$~univ. \\
    CoP decrease    & $\theta_c$ & $-$10.0\,px & $-$3.5\,px
      & \xmark~mat.-dep. \\
    Protection limit & $F_{\mathrm{lim}}$ & 45 & 100
      & \xmark~mat.-dep. \\
    Grasp-stop      & $F_{\mathrm{stop}}$ & 1.3 & 5.5
      & \xmark~mat.-dep. \\
    Weight exponent & $\gamma$ & 1.5 & 1.5
      & \cmark~univ. \\
    EMA coefficient & $\alpha$ & 0.3 & 0.3
      & \cmark~univ. \\
    \bottomrule
  \end{tabular}
\end{table}

%% file: experiments.tex
\section{EXPERIMENTS}
\label{sec:exp}

We evaluate TactileReflex through three experimental phases of
increasing complexity:
Phase~1 calibrates all thresholds and validates the three proxy
signals on two materials;
Phase~2 ablates each reflex channel to quantify its marginal
contribution;
and Phase~3 tests the full controller during a dynamic pouring task.

We use the thresholds derived from Phase 1 calibration:
slip threshold $\theta_s = 0.20$ (universal),
quiet-shear threshold $\theta_q = 0.12$/0.10 (soft/hard cup),
force protection limit $F_{\mathrm{lim}} = 45$/100 (soft/hard cup),
and CoP release threshold $\theta_c = {-10.0}/{-3.5}$\,px (soft/hard cup).
See Table~\ref{tab:thresholds} in Sec.~\ref{sec:calib} for complete parameter values.

\subsection{Phase 1: Threshold Calibration and Signal Validation}
\label{sec:exp_phase1}

Phase~1 executes the noise-statistics-based calibration protocol
described in Sec.~\ref{sec:calib}, deriving all controller thresholds
from the sensor's intrinsic noise profile on two contrasting materials.

\subsubsection{Experimental Protocol}
We use two contrasting cups:
(i)~a \textbf{soft cup} (disposable plastic, 3.5--5\,g,
0.3--0.5\,mm wall, highly compliant with narrow safe-force margin)
and (ii)~a \textbf{hard cup} (rigid 100\,mL measuring cup, negligible
deformation).
The full trial set comprises 19 trials (${\sim}$20\,000 frames)
across both materials in three progressively complex experiments:
\textbf{Exp.\,1} (static holding) establishes the noise baseline;
\textbf{Exp.\,2} (liftoff, support $\to$ unsupported) produces
slip events;
and \textbf{Exp.\,3} (manual push + press-release) produces shear
and load-decrease events.

\subsubsection{Signal Characterization and Threshold Derivation}
\label{sec:exp_e1}
\textbf{}\;

\textbf{Exp.\,1: Material-insensitive noise floor.}\;
Static holding trials (6\,477~frames across both materials) establish
the noise floor of each proxy.
$S_y$ noise $P_{95}$ is 0.08--0.11 (soft) and ${\sim}$0.08 (hard), nearly
identical despite 3--4$\times$ differences in $F_n$, confirming that
the median optical-flow statistic is material-insensitive.
This supports $\theta_q{=}0.12$ (soft) / 0.10 (hard).
$F_n$ exhibits strong L/R asymmetry (up to 10$\times$) reflecting
contact geometry; $P_{99.9}$ yields $F_{\mathrm{lim}}{=}45$ (soft)
and $100$ (hard).

\begin{figure}[t]
  \centering
  \includegraphics[width=\linewidth]{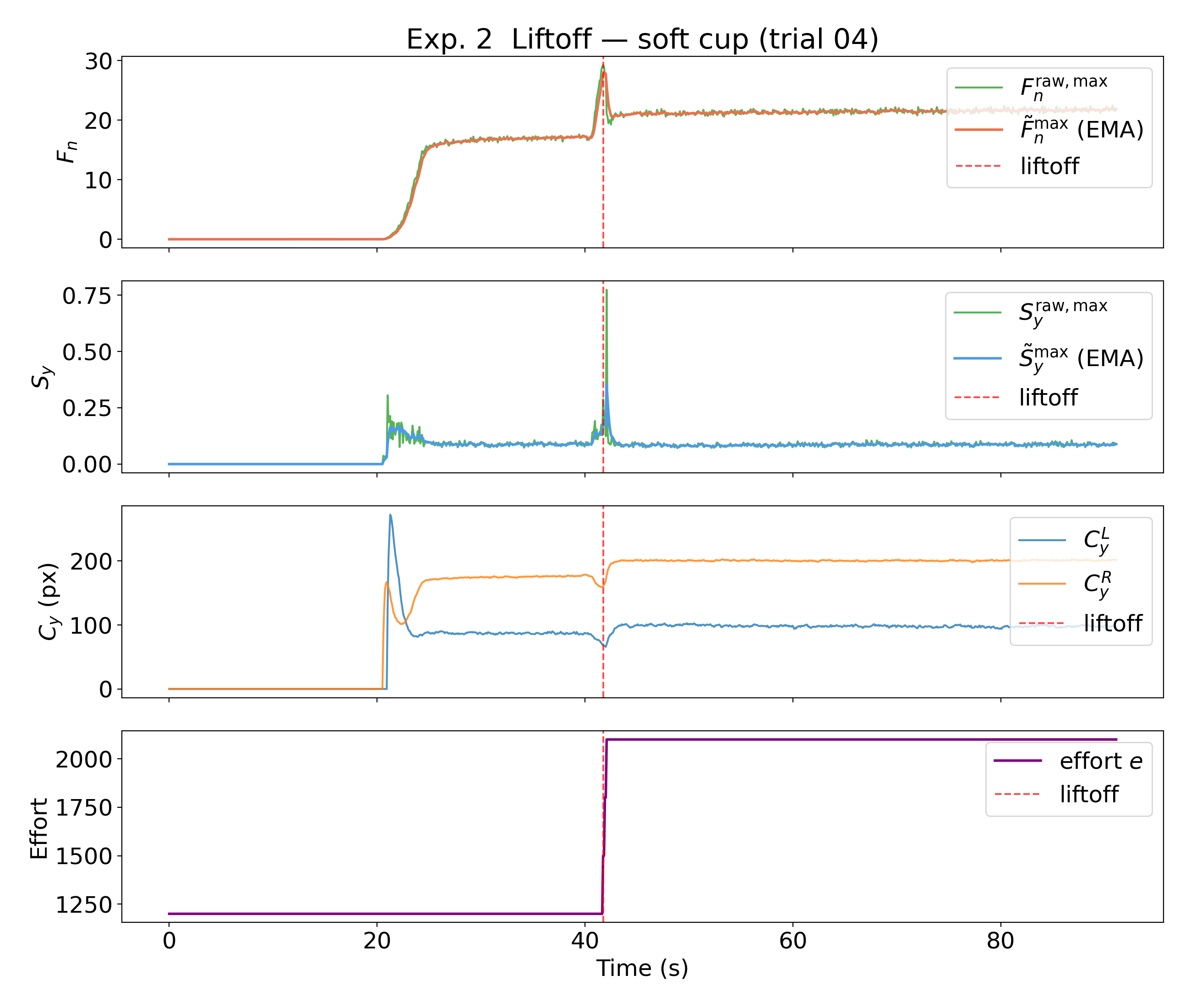}
  \caption{Exp.\,2 liftoff trial (soft cup).
  Upon liftoff, $S_y$ spikes above $\theta_s$, triggering three
  successive anti-slip responses that raise effort from
  1\,200 to 2\,100 within ${\sim}$0.34\,s.}
  \label{fig:liftoff_soft}
\end{figure}

\textbf{Exp.\,2: Universal slip threshold (100\% TPR, 0\% FPR).}\;\label{sec:exp_e2}
Removing the support block transfers the object weight to the gripper,
producing a transient $S_y$ spike.
Across both materials (7~trials, 18~slip events), $S_y$ peaks range
from 0.204 to 0.77, yielding 2.4$\times$ separation
(minimum peak / noise $P_{95}$; Table~\ref{tab:sy_separation})
to 6.9$\times$ at maximum peak.
The anti-slip reflex responded within ${\sim}$170--340\,ms
in every trial, preventing object drop
(Fig.~\ref{fig:liftoff_soft}).
We set the slip threshold by a conservative rule:
$P_{99.9}(S_y^{\mathrm{noise}}) \;\le\; \theta_s \;\le\; S_y^{\min\text{-}\mathrm{slip}}$,
where $S_y^{\mathrm{noise}}$ is computed from Exp.\,1 static-hold frames.
The threshold is placed just above the no-motion noise ceiling while
still at or below the weakest observed slip across both materials,
ensuring 0\% FPR and 100\% TPR simultaneously.
In our data, $S_y^{\min\text{-}\mathrm{slip}}=0.2006$ (a minimal hard-cup push) and $P_{99.9}(S_y^{\mathrm{noise}})<0.20$, so we use $\theta_s=0.20$.
Across ${\ge}100$ slip events and ${\sim}$6\,500 static frames on both materials,
this threshold yields 100\% TPR and 0\% FPR with the $\ge$ criterion in Eq.~\eqref{eq:slip_detect}.

\begin{table}[t]
\vspace{0.2cm}
  \centering
  \caption{$S_y$ noise--slip separation. $\theta_s{=}0.20$: 100\% TPR, 0\% FPR.
SNR denotes the separation ratio (slip min / noise $P_{95}$).}
  \label{tab:sy_separation}
  \begin{tabular}{lcccc}
    \toprule
    Source & Noise $P_{95}$ & Slip (min) & SNR & $N$ \\
    \midrule
    Soft Exp.\,1      & 0.112 & ---  & --- & --- \\
    Soft liftoff      & ---   & 0.429 & 3.8$\times$ & 11 \\
    Soft push  & ---   & 0.207 & --- & $\ge$55 \\
    \midrule
    Hard Exp.\,1      & 0.083 & ---  & --- & --- \\
    Hard liftoff      & ---   & 0.204 & 2.4$\times$ & 7 \\
    Hard push  & ---   & \textbf{0.201} & --- & 45 \\
    \bottomrule
  \end{tabular}
\end{table}

\textbf{Exp.\,3: $\theta_s$ cross-validated; $\theta_c$ per-material.}\;\label{sec:exp_e3}
Manual pushes add 45~slip events on the hard cup
($S_y$ peaks 0.201--1.45, all $\ge \theta_s$).
Press-release trials test the CoP channel: $\Delta C_y$ clusters
at $-3.0$ to $-3.6$\,px on the hard cup and $-10$ to $-29$\,px on
the soft cup, motivating $\theta_c{=}{-3.5}$ and $-10.0$\,px respectively.
The 3--8$\times$ difference in $|\theta_c|$ confirms that $\theta_c$
requires per-material calibration while $\theta_s$ does not.
\subsubsection{Cross-Material Comparison and L/R Asymmetry}
\label{sec:exp_crossmat}

Table~\ref{tab:cross_material} summarizes the cross-material and
L/R sensor asymmetry findings from Phase~1.

\begin{table}[t]
\vspace{0.2cm}
  \centering
  \footnotesize
  \caption{Cross-material comparison and L/R asymmetry robustness.
  $S_y$ is material-insensitive and area-invariant despite up to
  10$\times$ L/R asymmetry; $F_n$, CoP require per-material
  $F_{\mathrm{lim}}$, $\theta_c$.}
  \label{tab:cross_material}
  \setlength{\tabcolsep}{3pt}
  \begin{tabular}{lcccl}
    \toprule
    Metric & Soft cup & Hard cup & Ratio & Note \\
    \midrule
    \multicolumn{5}{l}{\textit{Noise (Exp.\,1 static holding)}} \\
    \quad $S_y\;P_{95}$
      & .08--.11 & .08--.08 & ${\sim}1\times$
      & \textbf{material-insens.} \\
    \quad $F_n$
      & 2--27 & 18--89 & 3--4$\times$
      & stiffness-sens. \\
    \midrule
    \multicolumn{5}{l}{\textit{Event response}} \\
    \quad Liftoff $S_y$ peak
      & 0.43--0.77 & 0.20--0.63 & ---
      & comparable \\
    \quad Release $|\Delta C_y|$
      & 10--29\,px & 3.0--3.6\,px & 3--8$\times$
      & \textbf{per-material $\theta_c$} \\
    \midrule
    \multicolumn{5}{l}{\textit{L/R sensor asymmetry (dominant / non-dominant)}} \\
    \quad $F_n$ L/R
      & 7--10$\times$ & 3--4$\times$ & --- & geometry-dep. \\
    \quad Contact area
      & 6--8$\times$ & 2--3$\times$ & --- & geometry-dep. \\
    \quad $S_y$ noise
      & ${\sim}1\times$ & ${\sim}1\times$ & ---
      & \textbf{area-invariant} \\
    \bottomrule
  \end{tabular}
\end{table}

L/R sensor asymmetry reaches 7--10$\times$ in $F_n$ and
6--8$\times$ in contact area on the soft cup, yet $S_y$ noise
remains ${\sim}1\times$ on both sides
(Eq.~\eqref{eq:sy_raw}), confirming that $\theta_s$ generalizes
across asymmetric contacts without adjustment.

\subsection{Phase 2: Ablation Study --- Three-Channel Necessity}
\label{sec:exp_ablation}

\subsubsection{Setup and Metrics}
To quantify the marginal contribution of each reflex channel,
we perform a systematic ablation with four controller configurations
(Table~\ref{tab:ablation}).
The soft cup is chosen because its narrow safe-force margin makes
deformation, rather than drop, the discriminating failure mode.
Each trial follows a standardized protocol:
grasp $\to$ liftoff $\to$ pull downward $\to$ press-release $\to$ push sideways.
We record drop (0/1) and irreversible deformation (0/1, defined as failure to self-recover upon release or creating spillage risk); success requires both = 0.
Metrics: $N_{\mathrm{slip}}$ (slip events), $F_n^{\mathrm{peak}}$ (peak contact intensity), $\Delta e$ (observed effort range).

\subsubsection{Results and Analysis}

Table~\ref{tab:ablation} summarizes the performance across 20 total trials 
($N{=}5$ per configuration).
All four configurations achieved 0\% drop rate (0/20 drops),
confirming that the base anti-slip channel reliably prevents object loss.
However, only the full three-channel configuration~D achieved
100\% success (5/5, no irreversible deformation).
Config~C (Sy+Release) achieved 1/5 in a trial where perturbation
magnitudes happened to remain low enough for \textsc{Release} to
recover effort despite the absence of \textsc{Protect};
configs~A and~B failed in every trial.

\begin{table}[t]
  \centering
  \caption{Ablation configurations and results (soft cup, $N{=}5$/config). Only D achieves 5/5 success.}
  \label{tab:ablation}
 
  \resizebox{\linewidth}{!}{
  \begin{tabular}{lccc | cccc}
    \toprule
    \textbf{Config} & \textbf{Slip} & \textbf{Rel.} & \textbf{Prot.} 
      & $N_{\mathrm{slip}}$ & $F_n^{\mathrm{peak}}$ 
      & $\Delta e$ & \textbf{Success} \\
    \midrule
    A: Sy-only    & \cmark & \xmark & \xmark 
      & $43.7{\pm}10.0$ & $63.0{\pm}12.6$ 
      & $3800{\pm}0$ & 0/5 \\
    B: Sy+Prot.   & \cmark & \xmark & \cmark 
      & $48.3{\pm}6.1$ & $70.4{\pm}17.2$ 
      & $1367{\pm}1457$ & 0/5 \\
    C: Sy+Rel.    & \cmark & \cmark & \xmark 
      & $49.7{\pm}14.6$ & $70.5{\pm}6.9$ 
      & $3600{\pm}346$ & 1/5 \\
    D: Full       & \cmark & \cmark & \cmark 
      & $40.3{\pm}4.7$ & $67.3{\pm}6.6$ 
      & $\mathbf{667{\pm}1155}$ & \textbf{5/5} \\
    \bottomrule
  \end{tabular}
  }
\end{table}

\begin{figure*}[t]
\vspace{0.2cm}
  \centering
  \includegraphics[width=\textwidth]{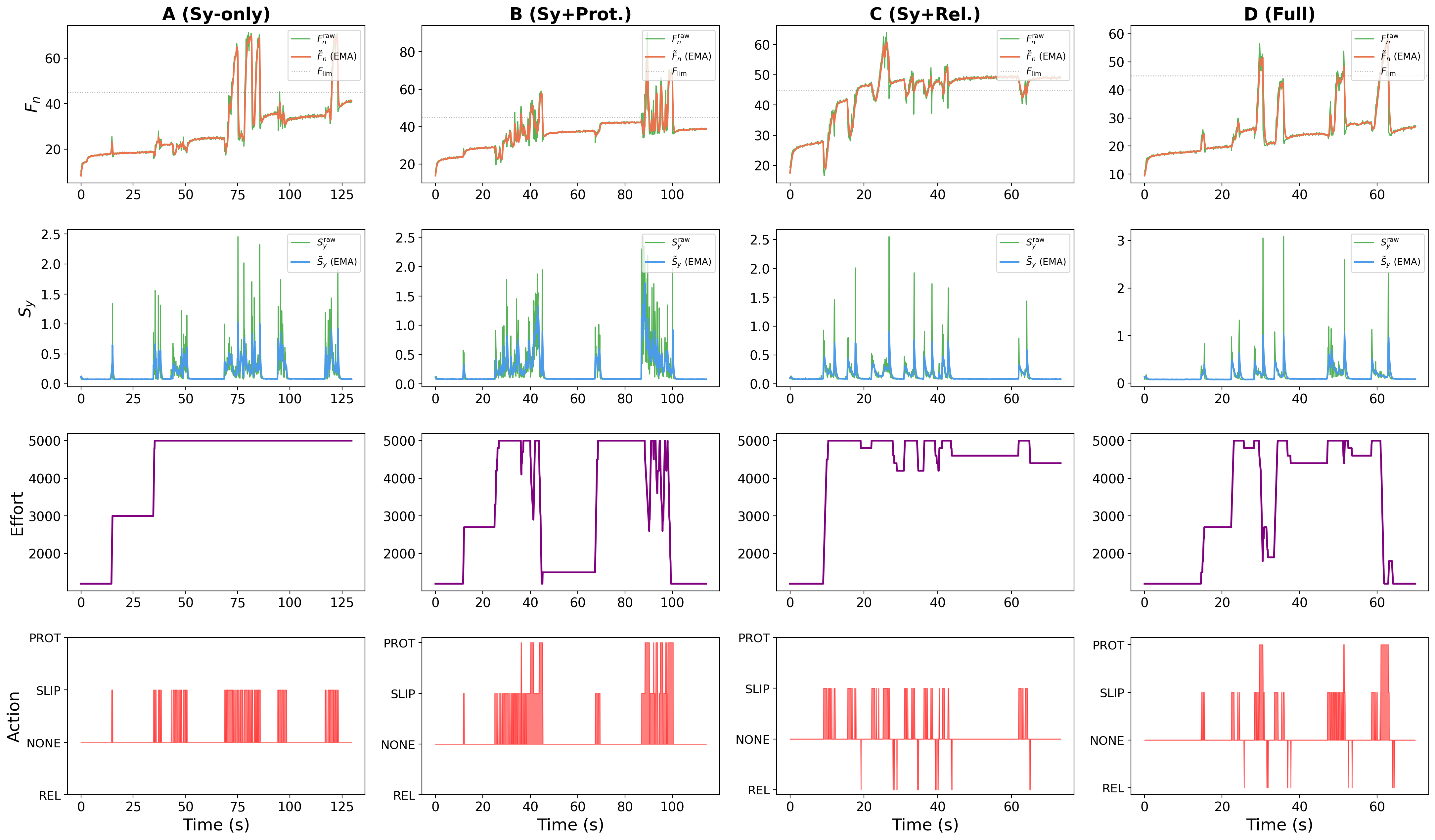}
  \caption{Ablation time-series (soft cup).
  A (Sy-only): effort saturates at $e_{\max}$.
  B (Sy+Protect): effort sawtooth oscillation.
  C (Sy+Release): release rarely fires.
  D (Full): all three channels active, effort recovers.}
  \label{fig:ablation_ts}
\end{figure*}

\textbf{Config~A (Sy-only): effort peaks and never decreases.}\;
Effort climbs monotonically to $e_{\max}{=}5\,000$ and remains
locked there for the rest of the trial ($\Delta e{=}3800$ in every
trial; Fig.~\ref{fig:ablation_ts}, column~A).
Without \textsc{Release} or \textsc{Protect}, no mechanism exists to
reduce the accumulated effort once the perturbation subsides, and the
cup irreversibly deforms under sustained over-gripping.

\textbf{Config~B (Sy+Protect): large-amplitude oscillation,
reactive only at $F_{\mathrm{lim}}$.}\;
\textsc{Protect} intervenes only \emph{after} $\tilde{F}_n^{\max}$
exceeds $F_{\mathrm{lim}}$, producing a characteristic effort
sawtooth with large swings ($+300$ per anti-slip step, $-400$ per
\textsc{Protect} step; Fig.~\ref{fig:ablation_ts}, column~B).
Because the correction engages 1--2~frames late and nothing drains
effort between spikes, $F_n$ peaks ($70.4{\pm}17.2$) repeatedly
overshoot the limit and still cause irreversible deformation.

\textbf{Config~C (Sy+Release): small, infrequent adjustment with
no force protection.}\;
\textsc{Release} only fires when shear is quiet
($\tilde{S}_y^{\max}{<}\theta_q$), yet at high effort residual shear
keeps this precondition blocked---\textsc{Release} triggers only $4.7$ times
vs.\ $9.3$ in~D, producing negligible effort reduction
($\Delta e{=}3600{\pm}346$;
Fig.~\ref{fig:ablation_ts}, column~C).
Without \textsc{Protect}, $F_n$ peaks go entirely unchecked:
there is no force ceiling to limit contact force, and the cup irreversibly deforms.

\textbf{Config~D (Full): timely adjustment with prompt force
protection.}\;
All three channels cooperate
(Fig.~\ref{fig:ablation_ts}, column~D):
when $F_n$ approaches $F_{\mathrm{lim}}$, \textsc{Protect} fires
promptly, capping force and reducing effort; the resulting drop in
$S_y$ satisfies the quiet-shear precondition, enabling
\textsc{Release} to detect CoP-based load decreases and further
drain effort.
Effort recovers to near-initial levels
($\Delta e{=}667{\pm}1155$), making this the only configuration
achieving 5/5 success.

\textbf{Summary.}\;
Each channel addresses a distinct failure mode:
anti-slip prevents drop (all configs achieve 0\% drop rate);
\textsc{Protect} caps force peaks but cannot \emph{actively} drain
accumulated effort (B still deforms);
\textsc{Release} can drain effort but needs the quiet conditions
that only \textsc{Protect} can create (C rarely fires).
Only their combination reliably prevents irreversible deformation
(Fisher exact test: config~D 5/5 vs.\ pooled A/B/C 1/15,
$p{<}0.001$).

\subsection{Phase 3: Dynamic Pouring Task}
\label{sec:exp_pour}

Pouring from a disposable plastic cup is a particularly demanding test:
the shifting liquid continuously redistributes the gravitational load,
leaving almost no static operating point.
We deliberately use the compliant soft cup to stress-test the reflex
layer under realistic conditions.

\subsubsection{Trajectory}
Trajectories via kinesthetic teaching (lift $\to$ tilt $\to$ pour $\to$ return).

\subsubsection{Experimental Conditions}
We test two water volume conditions: \textbf{Low (45\,ml, ${\sim}$1/3 cup)} and \textbf{High (90\,ml, ${\sim}$3/4 cup)}, with 5 trials per configuration.
Both groups share an \emph{identical grasping phase}: the reflex
controller closes the gripper and stabilizes in \textsc{Holding}
for 2\,s, after which the steady-state effort $e_{\mathrm{grasp}}$
and position $p_{\mathrm{grasp}}$ are recorded.
The two conditions then diverge:
\begin{itemize}
\item \textbf{No reflex} (baseline): replay $\mathbf{q}(t)$ with
      gripper frozen at $(e_{\mathrm{grasp}}, p_{\mathrm{grasp}})$.
      Tactile signals are still logged but no gripper adjustments
      are made;
\item \textbf{With reflex} (config~D): replay the same
      $\mathbf{q}(t)$ while TactileReflex adjusts effort and position
      in real time.
\end{itemize}
This design ensures the baseline effort is not an arbitrary constant
but the controller's own optimal static-hold output for each specific
cup and water volume, providing the fairest possible comparison.

\subsubsection{Metrics}
From the execution phase of each trial we extract:
\begin{itemize}
\item \textbf{Pour success}: water enters the target container
      \emph{and} the cup is neither dropped nor irreversibly deformed (binary, per trial);
\item \textbf{Slip fraction}: percentage of execution-phase frames
      in which $S_y^{\mathrm{raw}} \ge \theta_s$ (for reflex-off trials
      these are logged as \texttt{FROZEN\_SLIP});
\item \textbf{Effort range} $\Delta e$:
      difference between observed maximum and minimum effort during the pour
      ($\Delta e{=}0$ for the frozen baseline);
\item \textbf{Peak shear} $S_y^{\mathrm{peak}}$ and
      \textbf{min $F_n^L$} (weaker side), indicating
      slip severity and proximity to contact loss.
\end{itemize}

\subsubsection{Results and Analysis}

Table~\ref{tab:pour_results} summarizes the pouring results.
The outcome is clear: \textbf{every no-reflex trial failed to pour
water} (0/10), while \textbf{the reflex controller achieved 9/10
success} (5/5 at 45\,ml, 4/5 at 90\,ml).

\begin{table}[t]
  \centering
  \footnotesize
  \caption{Pouring results: no reflex (frozen effort) vs.\ reflex (config~D).
  Values are mean\,$\pm$\,s.d.\ across trials.
  Min $F_n^L$ is omitted for the reflex-on condition as the controller
  maintains continuous contact on both sensors throughout.}
  \label{tab:pour_results}
  \setlength{\tabcolsep}{3pt}
  \begin{tabular}{llccccc}
    \toprule
    Condition & Reflex
      & Success
      & Slip\,\%
      & $\Delta e$
      & $S_y^{\mathrm{peak}}$
      & min\,$F_n^L$ \\
    \midrule
    \multirow{2}{*}{\shortstack[l]{Low\\(45\,ml)}}
      & OFF & \textbf{0/5} & $42{\pm}4$\,\%
        & 0 & $2.5{\pm}0.5$ & $0.18{\pm}0.07$ \\
      & ON  & \textbf{5/5} & $35{\pm}4$\,\%
        & $3700{\pm}141$ & $1.8{\pm}0.4$ & --- \\
    \midrule
    \multirow{2}{*}{\shortstack[l]{High\\(90\,ml)}}
      & OFF & \textbf{0/5} & $63{\pm}1$\,\%
        & 0 & $3.4{\pm}0.7$ & $0.33{\pm}0.06$ \\
      & ON  & \textbf{4/5} & $42{\pm}5$\,\%
        & $3725{\pm}150$ & $3.3{\pm}0.7$ & --- \\
    \bottomrule
  \end{tabular}
\end{table}

\begin{figure}[t]
\vspace{0.2cm}
  \centering
  \includegraphics[width=0.47\linewidth]{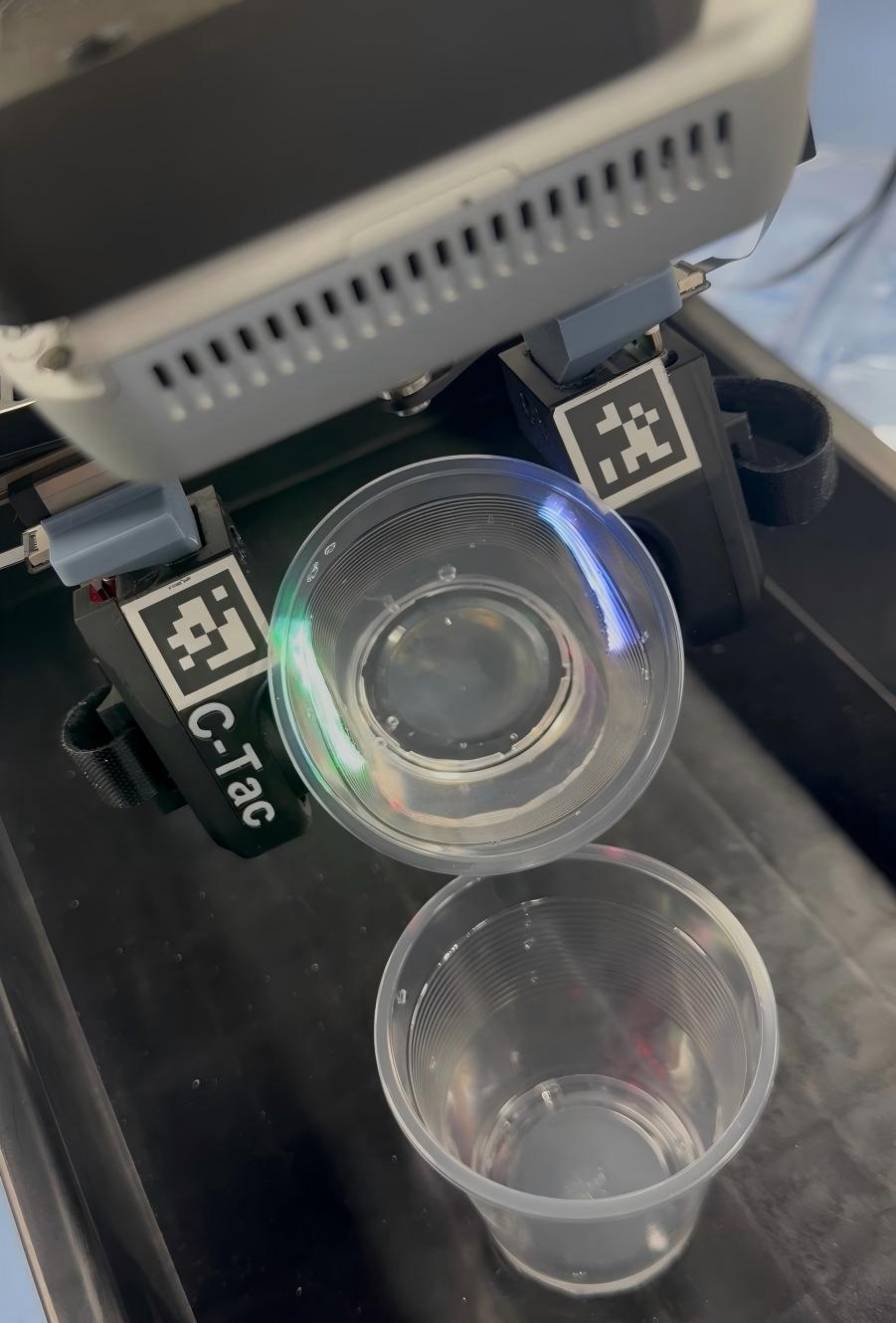}%
  \hfill
  \includegraphics[width=0.47\linewidth]{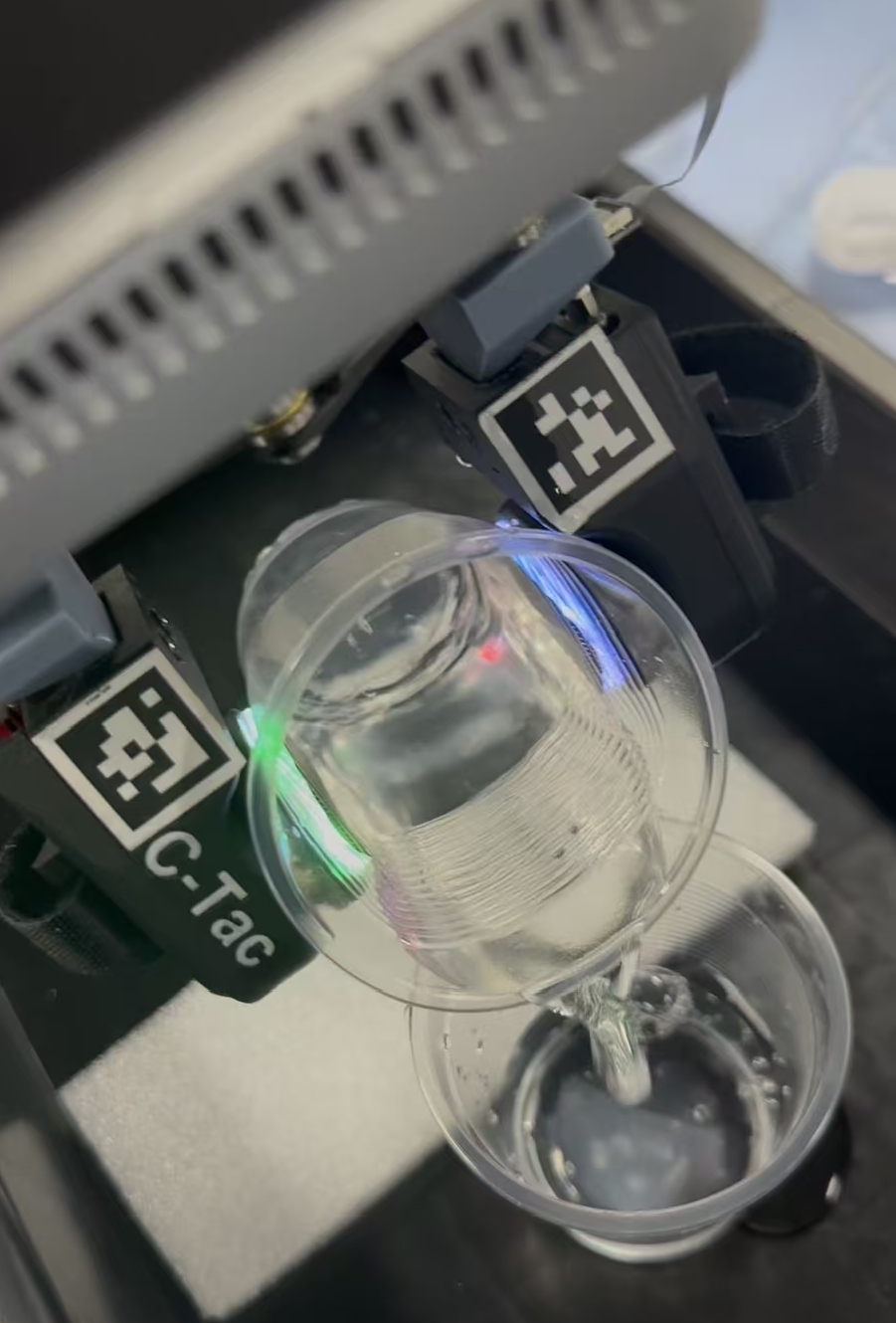}
  \caption{Pouring task comparison (45\,ml, slow).
  \textbf{Left:} without reflex, the cup continuously slid within
  the gripper during arm motion, drifting to an orientation where the
  cup mouth no longer faces the target container; water cannot be
  poured despite the arm reaching the correct tilt angle.
  \textbf{Right:} with reflex, anti-slip maintains the cup's
  orientation throughout the tilting motion; water is successfully
  poured into the container.}
  \label{fig:pour_photo}
\end{figure}

\textbf{Failure mode: pose drift, not drop.}\;
The cup was never dropped in any no-reflex trial.
Instead, with frozen effort, the cup \emph{continuously slid within the gripper},
progressively tilting until its mouth no longer aligned with the target
(Fig.~\ref{fig:pour_photo}, left).
This pose-drift failure is arguably more representative of real-world
manipulation than catastrophic drop, because the object is still
``held'' yet the task goal cannot be achieved.

\textbf{Slip fraction scales with water volume.}\;
Frozen-effort slip fraction rose from $42{\pm}4\%$ (45\,ml) to
$63{\pm}1\%$ (90\,ml), because heavier water increases the gravitational
torque during tilting.
At 90\,ml, L/R force asymmetry exceeded $90\times$, indicating
the cup was nearly held by one fingertip alone.

\textbf{Reflex suppresses slip and maintains pose.}\;
With the reflex active, the controller increased effort from the
frozen baseline (${\sim}$1\,200) to 4\,200--5\,000 and tightened
the grip by ${\sim}$3\,mm over the course of the trajectory.
Slip fraction dropped from 42\% to 35\% at 45\,ml and from 63\% to
42\% at 90\,ml.
Peak shear ($S_y^{\mathrm{peak}}$) was also reduced at 45\,ml
(from 2.5 to 1.8), indicating that the closed-loop response actively
dampened slip transients rather than merely tolerating them.
At 90\,ml, peak shear remained similar between groups (3.4 vs.\ 3.3)
because the gravitational torque is too large to fully suppress,
but the reflex's continuous effort adaptation kept the cup oriented
correctly in 4 of 5 trials (Fig.~\ref{fig:pour_photo}, right).
The single failure occurred when the cup's initial placement was
slightly off-center, causing an early slip cascade that exceeded
the controller's effort headroom ($e_{\max}{=}5\,000$).

\textbf{Three-channel synergy.}\;
At 90\,ml, one reflex-on trial triggered \textsc{Protect} for the first time in Phase~3
($F_n^{\mathrm{peak}}{=}57.9 > F_{\mathrm{lim}}{=}45$), demonstrating all three channels
operating in concert during a real dynamic task, consistent with the synergy identified in
Phase~2 (Sec.~\ref{sec:exp_ablation}).

\textbf{Frozen effort is statically optimal but dynamically insufficient.}\;
The frozen-effort baseline uses the controller's own steady-state output, not an arbitrary constant;
its universal failure (0/10; Fisher exact test vs.\ reflex 9/10: $p{<}0.001$) shows that
tilting induces time-varying gravitational torque requiring real-time closed-loop adaptation.

%% file: discussion.tex
\section{DISCUSSION AND CONCLUSION}
\label{sec:disc_concl}
We proposed a \emph{noise-statistics-based closed-loop reflex control paradigm} for vision-tactile sensing and instantiated it as TactileReflex. By automatically deriving all thresholds from intrinsic sensor noise via a brief 2-minute calibration, it eliminates manual tuning, external force calibration, and task-specific training. Operating at ${\sim}$12\,Hz, this prioritized three-channel controller (anti-slip, release, force-protect) achieves fast, stable grasping. Experiments confirm the cross-material universality of our area-invariant slip metric ($S_y$) and the strict necessity of all three channels operating in synergy to prevent irreversible deformation during dynamic tasks, where fixed-effort baselines categorically fail. As a plug-and-play safety layer, TactileReflex integrates seamlessly beneath fixed trajectories, learned policies, and VR teleoperation. By locally enforcing safety bounds, it enables scalable, haptic-free demonstration collection for fragile objects.

\textbf{Limitations and Future Work.}
Limitations include a gripper-centric scope and grayscale-only processing (chosen for lighting robustness and efficiency, but forgoing RGB information). Future work will exploit full-color tactile images, explore broader material sets---leveraging the self-calibrating thresholds---extend to arm-level reflexes, and integrate directly with VLA policies~\cite{brohan2023rt2,black2024pi0}.

%% file: reference.bib
@article{yuan2017gelsight,
  title = {{GelSight}: High-Resolution Robot Tactile Sensors for Estimating Geometry and Force},
  author = {Yuan, Wenzhen and Dong, Siyuan and Adelson, Edward H.},
  journal = {Sensors},
  volume = {17},
  number = {12},
  pages = {2762},
  year = {2017},
}

@article{lambeta2020digit,
  title = {{DIGIT}: A Novel Design for a Low-Cost Compact High-Resolution Tactile Sensor With Application to In-Hand Manipulation},
  author = {Lambeta, Mike and Chou, Po-Wei and Tian, Stephen and Yang, Brian and Maloon, Benjamin and Most, Victoria Rose and Stroud, Dave and Santos, Raymond and Byagowi, Ahmad and Kammerer, Gregg and others},
  journal = {IEEE Robotics and Automation Letters},
  volume = {5},
  number = {3},
  pages = {3838--3845},
  year = {2020},
}

@inproceedings{donlon2018gelslim,
  title = {{GelSlim}: A High-Resolution, Compact, Robust, and Calibrated Tactile-sensing Finger},
  author = {Donlon, Elliott and Dong, Siyuan and Liu, Melody and Li, Jianhua and Adelson, Edward and Rodriguez, Alberto},
  booktitle = {Proc. IEEE/RSJ Int. Conf. Intell. Robots Syst. (IROS)},
  pages = {1927--1934},
  year = {2018},
}

@inproceedings{ren2023mctac,
  title = {{MC-Tac}: Modular Camera-Based Tactile Sensor for Robot Gripper},
  author = {Ren, Jieji and Zou, Jiang and Gu, Guoying},
  booktitle = {Int. Conf. Intell. Robot. Appl. (ICIRA)},
  pages = {169--179},
  year = {2023},
}

@inproceedings{dong2019gelsight_slip,
  title = {Maintaining Grasps within Slipping Bounds by Monitoring Incipient Slip},
  author = {Dong, Siyuan and Ma, Daolin and Donlon, Elliott and Rodriguez, Alberto},
  booktitle = {Proc. IEEE Int. Conf. Robot. Autom. (ICRA)},
  pages = {3818--3824},
  year = {2019},
}

@article{james2018slip,
  title = {Slip Detection With a Biomimetic Tactile Sensor},
  author = {James, Jeremy W. and Pestell, Nathan and Lepora, Nathan F.},
  journal = {IEEE Robotics and Automation Letters},
  volume = {3},
  number = {4},
  pages = {3340--3346},
  year = {2018},
}

@inproceedings{su2015force,
  title        = {Force Estimation and Slip Detection/Classification for Grip Control Using a Biomimetic Tactile Sensor},
  author       = {Su, Zhe and Hausman, Karol and Chebotar, Yevgen and Molchanov, Artem and Loeb, Gerald E. and Sukhatme, Gaurav S. and Schaal, Stefan},
  booktitle    = {Proc. IEEE-RAS Int. Conf. Humanoid Robots},
  pages        = {297--303},
  year         = {2015},
  doi          = {10.1109/HUMANOIDS.2015.7363558}
}

@inproceedings{li2018slip,
  title = {Slip Detection with Combined Tactile and Visual Information},
  author = {Li, Jianhua and Dong, Siyuan and Adelson, Edward},
  booktitle = {Proc. IEEE Int. Conf. Robot. Autom. (ICRA)},
  pages = {7772--7777},
  year = {2018},
}

@inproceedings{li2014localization,
  title = {Localization and Manipulation of Small Parts Using {GelSight} Tactile Sensing},
  author = {Li, Rui and Platt, Robert and Yuan, Wenzhen and ten Pas, Andreas and Roscup, Nathan and Srinivasan, Mandayam A. and Adelson, Edward},
  booktitle = {Proc. IEEE/RSJ Int. Conf. Intell. Robots Syst. (IROS)},
  pages = {3988--3993},
  year = {2014},
}

@article{bauza2023tac2pose,
  title = {Tac2Pose: Tactile Object Pose Estimation from the First Touch},
  author = {Bauza, Maria and Bronars, Antonia and Rodriguez, Alberto},
  journal = {The International Journal of Robotics Research},
  volume = {42},
  number = {13},
  pages = {1185--1209},
  year = {2023},
}

@article{calandra2018more,
  title = {More Than a Feeling: Learning to Grasp and Regrasp Using Vision and Touch},
  author = {Calandra, Roberto and Owens, Andrew and Jayaraman, Dinesh and Lin, Justin and Yuan, Wenzhen and Malik, Jitendra and Adelson, Edward H. and Levine, Sergey},
  journal = {IEEE Robotics and Automation Letters},
  volume = {3},
  number = {4},
  pages = {3300--3307},
  year = {2018},
}

@misc{calandra2017feeling,
  title         = {The Feeling of Success: Does Touch Sensing Help Predict Grasp Outcomes?},
  author        = {Calandra, Roberto and Owens, Andrew and Upadhyaya, Manu and Yuan, Wenzhen and Lin, Justin and Adelson, Edward H. and Levine, Sergey},
  year          = {2017},
  note          = {arXiv:1710.05512},
}

@inproceedings{veiga2015stabilizing,
  title = {Stabilizing Novel Objects by Learning to Predict Tactile Slip},
  author = {Veiga, Filipe and van Hoof, Herke and Peters, Jan and Hermans, Tucker},
  booktitle = {Proc. IEEE/RSJ Int. Conf. Intell. Robots Syst. (IROS)},
  pages = {5065--5072},
  year = {2015},
}

@inproceedings{hogan2018tactile,
  title = {Tactile Regrasp: Grasp Adjustments via Simulated Tactile Transformations},
  author = {Hogan, Francois R. and Bauza, Maria and Canal, Oleguer and Donlon, Elliott and Rodriguez, Alberto},
  booktitle = {Proc. IEEE/RSJ Int. Conf. Intell. Robots Syst. (IROS)},
  pages = {2963--2970},
  year = {2018},
}

@article{romano2011human,
  title = {Human-Inspired Robotic Grasp Control With Tactile Sensing},
  author = {Romano, Joseph M. and Hsiao, Kaijen and Niemeyer, G{\"u}nter and Chitta, Sachin and Kuchenbecker, Katherine J.},
  journal = {IEEE Transactions on Robotics},
  volume = {27},
  number = {6},
  pages = {1067--1079},
  year = {2011},
}

@inproceedings{dafle2014extrinsic,
  title = {Extrinsic Dexterity: In-Hand Manipulation with External Forces},
  author       = {Dafle, Nikhil Chavan and Rodriguez, Alberto and Paolini, Robert and Tang, Bowei and Srinivasa, Siddhartha S. and Erdmann, Michael and Mason, Matthew T. and Lundberg, Ivan and Staab, Harald and Fuhlbrigge, Thomas},
  booktitle = {Proc. IEEE Int. Conf. Robot. Autom. (ICRA)},
  pages = {1578--1585},
  year = {2014},
}

@inproceedings{seita2021learning_deformable,
  title = {Learning to Rearrange Deformable Cables, Fabrics, and Bags with Goal-Conditioned Transporter Networks},
  author = {Seita, Daniel and Florence, Pete and Tompson, Jonathan and Coumans, Erwin and Sindhwani, Vikas and Goldberg, Ken and Zeng, Andy},
  booktitle = {Proc. IEEE Int. Conf. Robot. Autom. (ICRA)},
  pages = {4568--4575},
  year = {2021},
  doi = {10.1109/ICRA48506.2021.9561391},
}

@inproceedings{lin2020softgym,
  title = {{SoftGym}: Benchmarking Deep Reinforcement Learning for Deformable Object Manipulation},
  author = {Lin, Xingyu and Wang, Yufei and Olkin, Jake and Held, David},
  booktitle = {Proc. Conf. Robot Learn. (CoRL)},
  pages = {432--448},
  year = {2021},
}

@article{johansson1984,
  title = {Roles of Glabrous Skin Receptors and Sensorimotor Memory in Automatic Control of Precision Grip when Lifting Rougher or More Slippery Objects},
  author = {Johansson, Roland S. and Westling, G{\"o}ran},
  journal = {Exp Brain Res},
  volume = {56},
  number = {3},
  pages = {550--564},
  year = {1984},
}

@misc{brohan2023rt2,
  title = {{RT-2}: Vision-Language-Action Models Transfer Web Knowledge to Robotic Control},
  author = {Brohan, Anthony and Brown, Noah and Carbajal, Justice and Chebotar, Yevgen and Chen, Xi and Choromanski, Krzysztof and others},
  year = {2023},
  note = {arXiv:2307.15818},
}

@misc{black2024pi0,
  title = {$\pi_0$: A Vision-Language-Action Flow Model for General Robot Control},
  author = {Black, Kevin and Brown, Noah and Driess, Danny and Esmail, Adnan and Equi, Michael and Finn, Chelsea and others},
  year = {2024},
  note = {arXiv:2410.24164},
}

@misc{tafvla2024,
  title = {Tactile-{VLA}: Unlocking Vision-Language-Action Model’s Physical Knowledge for Tactile Generalization},
  author = {Huang, Jialei and Wang, Shixian and Lin, Feng and Hu, Yang and Wen, Chen and Gao, Yingjie},
  year = {2025},
  note = {arXiv:2507.09160},
}

@article{liu2024safety_layer,
  title = {Safe Reinforcement Learning on the Constraint Manifold: Theory and Applications},
  author = {Liu, Puze and Bou-Ammar, Haitham and Peters, Jan and Tateo, Davide},
  journal = {IEEE Transactions on Robotics},
  volume = {41},
  pages = {3442--3461},
  year = {2025},
}

@inproceedings{farneback2003,
  title = {Two-Frame Motion Estimation Based on Polynomial Expansion},
  author = {Farneb{\"a}ck, Gunnar},
  booktitle = {Image Analysis},
  pages = {363--370},
  year = {2003},
}

@manual{ati_ft_faq2020,
  title = {FAQ -- Force/Torque Sensors},
  organization = {ATI Industrial Automation},
  year = {2020},
}

@article{chavez2019insitu,
  title = {Six-Axis Force Torque Sensor Model-Based In Situ Calibration Method and Its Impact in Floating-Based Robot Dynamic Performance},
  author = {Andrade Chavez, Francisco Javier and Traversaro, Silvio and Pucci, Daniele},
  journal = {Sensors},
  volume = {19},
  number = {24},
  pages = {5521},
  year = {2019},
}
